\documentclass[11pt]{article}
\usepackage{acl2014}
\usepackage[utf8]{inputenc}
\usepackage{times}
\usepackage{url}
\usepackage{latexsym}
\usepackage{amsmath}
\usepackage{paralist}
\usepackage{pdfpages}
\DeclareMathSizes{11}{9}{13}{9}
\title{Gap-weighted subsequences for automatic cognate identification and
phylogenetic inference}
\author{Taraka Rama \\ Språkbanken \\ University of Göteborg}
\date{}

\begin{document}
\maketitle
\begin{abstract}
In this paper, we describe the problem of cognate identification and its
relation to phylogenetic inference. We introduce subsequence based features for
discriminating cognates from non-cognates. We show that subsequence based
features perform better than the state-of-the-art string similarity measures for
the purpose of cognate identification. We use the cognate judgments for the
purpose of phylogenetic inference and observe that these classifiers infer a
tree which is close to the gold standard tree. The contribution of this paper
is the use of subsequence features for cognate identification and to employ the
cognate judgments for phylogenetic inference.
\end{abstract}

\section{Introduction}\label{sec:intro}
Historical linguistics, the oldest branch of modern linguistics, studies how
languages change and attempts to infer the genetic relationship between
languages with suspected relationship. In this context, genetic relationship
means
that two languages are solely similar due to their descent from a common
ancestor and not due to structural similarity. Identification of cognates is a
very important step prior to the positing of any genetic relationship between
two languages.

Cognates are words across languages whose origin can be traced
back to a common ancestor. For example, English $\sim$ German \emph{night}
$\sim$ \emph{Nacht} `night' and English \emph{hound} and German \emph{Hund}
`dog' are cognates whose origin can be traced back to a common ancestor. In
historical linguistics, cognates are identified through the application of
the comparative method~\cite{rankincomparative}. Sometimes, cognates are not
revealingly similar but have changed substantially over time such that they do
not share form similarity. An example of such a cognate pair is the English
\emph{wheel} and Sanskrit \emph{chakra} `wheel', which can be traced back
to Proto-Indo-European (PIE) *$k^wek^welo$.
% The synchronic words can be derived from the PIE
% word through the application of series of sound laws.

When historical linguists work with the comparative method, they compare
basic vocabulary, phonological correspondences, grammatical forms,
and morphological paradigms to establish relationship between languages
suspected of common descent. However, performing a large scale automatic
grammatical correspondence analysis presupposes that we have well-defined
morphological analyzers for ancient, extinct, and under-documented languages.
% As emphasized by
% \newcite{meillet1967comparative}, grammatical and phonological evidence
% outweigh evidence from basic vocabulary comparison. For instance, Hittite (an
% extinct
% language earlier spoken in modern-day Turkey) was shown to belong to the
% Indo-European
% family largely
% based on morphological evidence~\cite{campbell2008language}.

Basic vocabulary lists such as the ones devised by Morris
Swadesh~\cite{swadesh1952lexico}, provide a suitable
testing
ground for applying machine learning algorithms to automatically identify
cognates. Some standardized word lists come with cognate information and,
subsequently, is used to infer the
relationship between languages under purview~\cite{dyen1992indoeuropean}. In the
related field of
computational biology, the term \emph{phylogenetic inference} is in vogue to
signify computational methods which infer relationship between biological
species~\cite{felsenstein2004inferring}. The same term has come to refer to the
identification of genetic relationships between languages
also~\cite{McMahon:05}.

\newcite{swadesh1952lexico} developed lexicostatistics as a technique to
infer relationships between languages. In this effort, Swadesh posited a list of
basic vocabulary items, ranging from sizes 100--200 that are supposed to be
universal, culture-free, and resistant to replacement over time. In positing
these word lists,
Swadesh intended to develop a concept list where the translational equivalents
for each
language would be provided by language experts. In the next step, the
cognacy
status
between a pair of words is determined through the application of the comparative
method. Finally, the similarity of a language pair is defined as the
total number of shared cognate word pairs divided by the total
number of word pairs. The pair-wise distance matrix
computed from this step can then be supplied to a clustering algorithm such as
UPGMA~\cite{sokal1958statistical}\footnote{Also known as average-linking
clustering in NLP~\cite{Manning:99}.} to infer a tree between the
languages.

Thus, the tasks of establishing relationship between languages
as well as the identification of cognates are closely related tasks where the
output of the latter serves as an input to the former. Automatic
cognate identification, as defined in computational linguistics literature,
refers to the application of string similarity or phonetic similarity
algorithms either independently, or in tandem with machine learning algorithms
for determining if a
given word pair is cognate or not~\cite{inkpen2005automatic}.\footnote{
In NLP, even borrowed words (\emph{loanwords}) which usually have strong
semantic as well as form similarity are referred to as cognates. In contrast,
historical linguistics makes a stark distinction between loanwords and cognates.
An example of a loanword is English \emph{beef} from Norman
French. \emph{correlates} \cite{McMahon:05} include cognates and
borrowings. In NLP, the words descending from an
ancestral language are referred to as `genetic
cognates'~\cite{kondrak2005cognates}. In this paper, we use cognates to refer to
those words whose origin can be traced back to a common ancestor.}

The approaches developed by \newcite{Kondrak:06} and
\newcite{inkpen2005automatic} supply different string distances between a pair
of words as features to a linear classifier. Usually, a linear classifier such
as Support Vector Machine (SVM) is trained with labeled positive (``cognates'')
and negative (``non-cognates'') examples and tested on a held-out dataset.
Cognate information has been applied to the tasks of sentence
alignment~\cite{simard1993using} and statistical machine
translation~\cite{kondrak2003cognates}.

In this paper, we use subsequence based features for automatic cognate
identification as well as phylogenetic inference. We show that subsequence
based features outperform word similarity measures for the task of cognate
identification. We
motivate the use of subsequence based features in terms of linguistic examples
and then proceed to formulate the subsequence based features that can be
derived from string kernels~\cite{shawe2004kernel} developed for text
categorization task~\cite{lodhi2002text}. In IR literature, string
subsequences go under the name of skip-grams~\cite{jarvelin2007s}.

The rest of the paper is structured as followed. In section~\ref{sec:probdef},
we define the two problems of automated cognate identification and
phylogenetic inference. We describe related work in section~\ref{sec:rel}.
Section~\ref{sec:cogid} describes subsequence features, experimental setup,
dataset, evaluation measures, and results. In section~\ref{sec:phyli}, we
describe our phylogenetic experiment setup and the evaluation measure for the
inferred tree. We discuss the results of our experiments as we present them.
Finally, we conclude and provide pointers to future direction in
section~\ref{sec:concl}.

\section{Two problems}\label{sec:probdef}
In this paper, we work with identifying cognates in Swadesh lists for the
Indo-European family. The Swadesh lists -- of length $200$ -- for $84$
Indo-European languages were compiled by \newcite{dyen1992indoeuropean}. As
mentioned before, the Swadesh lists contain the lexical realization for a
concept and its cognate class. A cognate class is a function mapping a
set of multiple items belonging to different languages to a unique cognate
class number (CCN).
Hence, for each concept, positive training instances consist of pairs of
words belonging to different languages that share a CCN. If the words in the
pair do not share a CCN number, then the word pair is labeled as a negative
instance. We intend to explore the efficacy of subsequence features to the
following problems:

\begin{compactenum}
 \item In a scenario where there are few positive examples and a very large
number of negative examples, how well do subsequence features perform over
string
similarity measures in the task of cognate identification?
\item In many families, the information about cognacy judgments is partially
available. In such a case, how well can a classifier trained on partial data
be used to identify cognates in the remaining languages? Can the classifier
generalize over the language family? Can the cognate judgments inferred
from the previous step be used to infer
a phylogenetic tree?
\end{compactenum}

\section{Related work}\label{sec:rel}
\newcite{ellison-kirby:2006:COLACL} use scaled edit distance (normalized by
average length)
to measure the intra-lexical divergence in a language. This step yields a
language-internal probability distribution. They then apply the KL-divergence
measure to calculate the distance between a language pair. This step is
repeated for all the $42\times 83$ language pairs from Dyen et al.,'s IE
database to yield a distance matrix.
The distance matrix is then used to infer a tree for the IE language.
Unfortunately, they perform a qualitative evaluation of the inferred tree and
do not compare the tree to the standard tree inferred by experts of the language
family. The authors mention string kernels but do not pursue this line of
research further.

% In another paper, \newcite{Atkinson:06} apply a cognate gain-loss model coupled
% with Bayesian MCMC for inferring the phylogenetic tree for IE family. They
% convert the cognacy judgments given in the Dyen et al.,'s IE database into a
% binary feature matrix such that each language is represented by a binary
% feature vector. Then they supply the binary vectors to a Bayesian tree
% inference program to yield a set of phylogenetic trees. They find that their
% consensus tree is close to the gold standard tree given by historical
% linguistics. It has to be pointed out that although their method searches over a
% large tree space, the method requires binary coded cognacy judgments
% before-hand.

\newcite{bouchardcote2013} employ a graphical model to reconstruct the
proto-word forms from the synchronic word-forms for the Austronesian language
family. They compare their automated reconstructions with the ones
reconstructed by historical linguists and find that their model beats an
edit-distance baseline. However, their model has a strict requirement
that the tree structure between the languages under study has to be known
before-hand.

\newcite{greenhill2011levenshtein} argues against the use of vanilla edit
distance for cognate identification and language distance computation. However,
a recent paper by \newcite{hauer-kondrak:2011:IJCNLP-2011}
shows that a combination of edit distance and other string similarity
measures, supplied as features to a SVM classifier, will boost
the cognate identification accuracy.

\newcite{hauer-kondrak:2011:IJCNLP-2011}\footnote{Henceforth, referred to as
HK.} supply different string similarity scores as features to a SVM
classifier for determining if a given word pair is a cognate or not. The authors
also employ an additional binary language-pair feature -- that is used
to weigh the language distance -- and find that the
additional feature assists the task of semantic clustering. In this task, the
cognacy judgments given by a linear classifier is used to flat cluster the
lexical items belonging to a single concept. The clustering quality is
evaluated against the gold standard cognacy judgments. Unfortunately, the
experiments of these scholars cannot be replicated since the partitioning
details of
their training and test datasets is not available.

In our experiments, we use edit distance as the sole feature for a baseline
classifier. We also compare our results with the results of the classifiers
trained from HK-based features.

\section{Cognate identification}\label{sec:cogid}
The vanilla edit distance measure counts the minimum number of insertions,
deletions, and substitutions required to transform a word into another word.
Identical words have $0$ edit distance. For example, the edit distance between
two cognates English \emph{hound} and German \emph{hund} is $1$. Similarly, the
edit
distance between Swedish \emph{i} and Russian \emph{v} `in', which are cognates,
is $1$. The edit distance treats both of the cognates at the same level
and
does not reflect the amount of change which has occurred in the Swedish and
Russian words from the PIE word.

Another string similarity measure such as Dice\footnote{In general, Dice
between two sets is defined as the ratio of number of shared elements to the
total number of elements in both the sets.} estimates word similarity as
the ratio between the number of common bigrams
to estimate the similarity between two words. The similarity between Lusatian
\emph{dolhi} and Czech \emph{dluhe} `long' is $0$ since they do not share any
common bigrams and the edit distance between the two strings is $3$. Although
the two words share all the consonants, the Dice score is $0$ due to the
intervening vowels.
% Add turchin's reference.

Another string similarity measure, Longest Common Subsequence (LCS) measures the
length of the longest common subsequence between
the two words. The LCS is $4$ (\emph{hund}), $0$, and $3$ (\emph{dlh}) for the
above
examples. One can parade a number of examples which are problematical for the
simple-minded
string similarity measures. Alternatively, string kernels in machine learning
research offer a way to exploit the similarities between two words without any
restrictions on the length and character similarity.

\subsection{Subsequence features}
Edit distance in its rawest form aligns two strings based on the minimum number
of edit operations. Edit distance neither makes any distinction between
aligning vowels to consonants nor does it account for the similarity between
two sounds (e.g., /p/ and /b/). Multiple approaches have been proposed to
alleviate
these shortcomings.
\newcite{wieling2009evaluating} propose a
Vowel-Consonant-constrained edit
distance, based on PMI (pair-wise mutual information), for the
purpose of extracting matching sounds between two words.\footnote{Vowels do not
align with consonants.\cite{prokic2010families}} They apply their method to
dialect data and find that their method identifies the traditional dialectal
boundaries. In extension,~\newcite{jager2013phylogenetic} used a PMI-based edit
distance on a training dataset to compute the distance between phonetic
symbols. The symbol similarity matrix was used to compute pair-wise language
distances. The pair-wise language distances were then compared to the gold
standard classification. They find that PMI-based edit distance outperforms
edit-distance based language distances.

\newcite{turchin2010analyzing} employ
matching consonant classes to determine the similarity between two words. These
approaches require explicit formalization of linguistic constraints depending on
the
languages under consideration. In fact, vowel quality is known to vary across
time. If we drop the vowels in the Czech-Lusatian word pair, then the words
are identical. In another study,~\newcite{list:2012:LINGVIS2012} uses a
permutation based method to learn the similarity between sounds and employs the
technique to cluster identified cognates for a concept. A SVM classifier learns
the weight for a subsequence feature and combines the learned weights of the
features without any human intervention.

Subsequences of length greater than $1$ also take context into account.
Subsequences as formulated below weigh the similarity between two words based
on the number of dropped characters and combine vowels and consonants
seamlessly. Having motivated why subsequences seems to be a good idea, we
formulate
subsequences below.

We follow the notation given in \newcite{shawe2004kernel} to formulate our
representation of a word (string). Given a string $s$, the subsequence vector
$\Phi(s)$
is defined as follows. The string $s$ can decomposed as $s_1,\dotsc,s_{|s|}$
where $|s|$ denotes the length of the string. Let $\overrightarrow{I}$ denote a
sequence of indices $(i_1,\dotsc,i_{|u|})$ where, $1 \le i_1 < \dotsc
< i_{|u|} \le |s|$. Then, a subsequence $u$ is a sequence of characters
$s[\overrightarrow{I}]$. Note that a subsequence can occur multiple times in a
string. Then, the weight of $u$, $\phi(u)$ is defined as
$\sum_{\overrightarrow{I}:u=s[\overrightarrow{I}]}
\lambda^{l(\overrightarrow{I})}$ where, $l(\overrightarrow{I}) = i_{|u|}-i_1+1$
and $\lambda \in [0, 1]$ is a decay factor. The subsequence vector $\Phi(s)$ is
$(\phi_{u_1}\dotsc \phi_{u_{|\Sigma ^*|}})$ where, $\Sigma ^* =
\bigcup_{n=0}^\infty \Sigma^n$ is the set of all strings from an alphabet
$\Sigma$. In our experiments, we fix the value of $\lambda$ at $0.5$.

The $\lambda$ factor is exponential and penalizes $u$ over long
gaps in a string. Due to the above formulation, the frequency of a
subsequence $u$ is also taken into account. In our experiments, we observed that
a few thousand word pairs did not have a single character in common. In such
a scenario, we default to class-based subsequence features by mapping a
$\Sigma$ in $u$ to its Consonant/Vowel class -- $\Sigma\mapsto\{C, V\}$. As a
preliminary step, we map each string $s$ into its $C, V$ sequence $s_{cv}$ and
then compute the subsequence weights.\footnote{$V=\{a,e,i,o,u,y\}$, $C=\Sigma
\setminus V$}

A combined subsequence vector $\Phi(s+s_{cv})$ is further normalized by its
norm, $\|\Phi(s+s_{cv})\|$, to transform into a unit vector. The common
subsequence vector $\Phi(s_1,s_2)$ is composed of all the common subsequences
between $s_1,s_2$. The weight of a common subsequence is
$\phi_{u}^{s_1}+\phi_{u}^{s_2}$.

% The subsequence
% vectors of two strings $s_1$ and $s_2$ are then used to output the subsequence
% (feature) vector for the string pair.

\newcite{moschitti2012modeling} list the features of the above weighting
scheme.
\begin{compactitem}
 \item Longer subsequences receive lower weights.
\item Characters can be omitted (called gaps).
\item The exponent of $\lambda$ penalizes recurring subsequences with
longer gaps.
\end{compactitem}
For a string of length $m$ and a pre-determined subsequence length $p$,
the computational complexity is in the order of $\mathcal{O}(mp)$.

% Write a equation about primal form and conversion into probability.
On a linguistic note, gaps are consistent with the prevalent
sound changes such as sound loss, sound gain, and
assimilation\footnote{A sound can assimilate to a neighboring sound. Sanskrit
\emph{agni} $>$ Prakrit \emph{aggi} `fire'. Compare the
Latin form \emph{ignis}.}, processes which alter word forms in an ancestral
language causing the daughter languages to have different surface forms.
The $\lambda$ factor weighs the number of gaps found in a subsequence.
For instance, the Sardinian word form for `fish' \emph{pissi} has the
subsequence \emph{ps} occurring twice but with different weights: $\lambda^3
, \lambda^4$.

The combined feature vector, for a word pair, is used to train a SVM classifier.
In our experiments, we use the LIBLINEAR package
\cite{fan2008liblinear} to solve the primal problem with L$_2$-regularization
and L$_2$-loss. The next subsection describes the makeup of the dataset. We use
the default parameters since we did not observe any difference in our
development experiments.

\subsection{Dataset}

We used the publicly available Indo-European
dataset~\cite{dyen1992indoeuropean} for our experiments. The dataset has
$16,520$ lexical items for $200$ concepts and $84$ language varieties. Each word
form is assigned to a unique CCN. A concept can have multiple word forms. In
such a case, we randomly pick one word and discard the rest of the forms. There
are more than $200$ identical non-cognate pairs in the dataset.

For the first experiment, we extracted all word pairs for a concept
and assigned a positive label if the word pair has an identical CCN; a negative
label, if the word pair has different CCNs. We extracted a total
of $674,192$ word pairs out of which $158,787$ are cognates.

The word length is an important parameter in our experiments since it
gives an
index of how far the value of subsequence length, $p$, should be tested. We
found that the average
word length is about $4.79$ and the median is $5$. There are about $928$ words
which have a word length greater than $7$. Hence, we tested the effect of $p$
from $1$ to $7$. We report the results for different values of $p$.

\begin{table}[htbp]
\centering
\small
\begin{tabular}{|l|c|}
\hline
Subfamily & \# of languages \\\hline
Germanic & $14$ \\
Indo-Iranian & $18$ \\
Romance & $14$ \\
Slavic & $13$ \\
Others & $25$\\\hline
\end{tabular}
\caption{Number of languages in each subfamily.}
\label{tab:data}
\end{table}

The second experiment involves cognate identification as a step
towards phylogenetic inference. In this experiment, we split the $84$ languages
into training and test sets based on their membership in subfamilies. The
IE dataset has $84$ languages belonging to $8$ different subfamilies. Out of
these, Germanic, Indo-Iranian, Romance, and Slavic have more than $10$ languages
(cf. table~\ref{tab:data}). The rest of the languages are distributed across
the Celtic, Baltic, Armenian, and Albanian groups.

\begin{table}[h!]
\centering
\small
\begin{tabular}{|l|cc|}
\hline
& positive ($+$ve) & negative ($-$ve) \\\hline
training & $38,722$ & $135,658$\\
test & $39,432$ & $119,389$\\\hline
\end{tabular}
\caption{Number of positive and negative examples in the training and
test datasets.}
\label{tab:train_test_data}
\end{table}

We merged all the groups with less than $10$ languages into a single
group of $25$ languages, ``Others''. Then, we randomly split each subfamily into
a training
and testing dataset of roughly equal size. Subsequently, we merged the
subfamilies' training datasets into a single training dataset. We
followed the same merging procedure with the test datasets to create a single
test dataset for the whole language family. Finally, we
extracted the subsequence feature vectors for each labeled word pair. The
details of dataset is given in
table~\ref{tab:train_test_data}. The idea behind this setup is explained in
question 2 under section~\ref{sec:probdef}.
% As a final note, we had to discard $235$ word pairs which had no symbol common
% in surface sequences or $CV$ sequences to compare against HK's features. It is
% interesting to note that $170$ of these word pairs (actually cognates) belong to
% the concept ``in''. A cursory analysis suggests that the word pairs belong to
% the Slavic and other IE languages respectively. We regard this example as a
% limitation to our method. The Swedish form \emph{i} and Russian \emph{v}
% descended from a common word which was clipped to retain a vowel in Swedish and
% a consonant in Russian.

\subsection{Evaluation measures}
In this section, we describe the different measures for evaluating the results
of our experiments. In our first experiment, the performance of the linear
classifier was evaluated using five-fold cross-validation accuracy. The
accuracy measure is defined as below:

\begin{subequations}
\small
\begin{align}
N = TN+TP+FN+FP \\
ACC = \frac{TP+TN}{N}
\end{align}
\end{subequations}
\normalsize
where, TP: number of true
positives, TN: true negatives, FP: false positives, FN: false negatives, and
$N$ shows the total number of test instances.

In the second experiment of cognate identification, we use Matthews Correlation
coefficient (MCC) and Average Precision (AP) for evaluating the performance of
our classifier. MCC~\cite{matthews1975comparison} is a comprehensive evaluation
measure which takes TP, TN, FP, and FN into account when computing the agreement
between the predicted binary vector, $\mathbf{\hat{y}}$ and the gold standard
binary vector, $\mathbf{y}$. The calculation of MCC is not straightforward and
is given in equations \ref{eq:mcc}a--\ref{eq:mcc}c. MCC $\in [-1, +1]$ where a
score of $-1$ suggests perfect disagreement whereas $+1$ suggests perfect
agreement. MCC is used when there is a difference in the size of the classes in
the test dataset. In our case, the number of negative examples are thrice the
number of positive examples. MCC is a special case of Pearson's $r$, which
measures the agreement between two binary vectors.

\begin{subequations}\label{eq:mcc}
\small
\begin{align}
 S & = \frac{TP+TN}{N}\\
P & = \frac{TP+FP}{N}\\
MCC & = \frac{TP/N - S\times P}{\sqrt{PS(1-P)(1-S)}}
\end{align}
\end{subequations}
\normalsize

The classification score given by a linear classifier for a test instance is
transformed into a probability score through sigmoid function. Thus a test
instance is labeled as positive if it has a probability score $>0.5$,
else is classified as negative. In a classic information retrieval task setting,
one would look at the precision, $p(r)$, plotted as a function of recall $r \in
[0,1]$, to observe the performance of the classifier for various classifier
thresholds. Hence, we report the average precision (AP) score, defined as
$\int_0^1 p(r) dr$, for each experimental setting.

It is worth noting that \newcite{kondrak-tal} employs
11-point interpolated average precision for evaluating different similarity
algorithms on a small test set consisting of five languages. In our
experiments, we use a larger test set of $41$ languages. The AP score also
measures the robustness of a classifier against different thresholds. If a
classifier ranks low at AP but evaluates well for other measures, it suggests
that the classifier is not robust to the shifting probability thresholds.

We use the three evaluation measures to check if the classifiers perform well on
the task of detecting TPs and TNs. Ideally, a cognate identifier system should
perform well on both positive and negative examples. The difference between MCC
and AP is that MCC evaluates the performance of a classifier on both positive
and negative examples for a fixed threshold.

\subsection{Results}
In this subsection, we describe the results of our experiments on cognate
identification using subsequence features. In the first experiment, we perform
a five-fold cross-validation on the labeled positive and negative examples.
Then we move to report our results on the combined feature vectors comprising
subsequence features and word pair similarity features. The following word pair
similarity features from \newcite{hauer-kondrak:2011:IJCNLP-2011} are used in
our experiments:
\begin{compactitem}
 \item Edit distance
\item Length of longest common prefix
\item Number of common bigrams
\item Lengths of individual words
\item Absolute difference between the lengths of the words
\end{compactitem}
which is referred to as HK in all the tables.

\subsubsection{Cross-validation experiments}
The main aim of this experiment is to determine if subsequence features work at
least as well as HK features on a dataset split into five folds. The accuracies
presented in table~\ref{tab:cv} show that subsequence features work better than
HK for all values of $p$. The highest accuracy is at $p=7$. Even the
subsequences of length $2$ outperform the HK-based classifier and baseline
classifier. In the rest of the paper, we do not use the baseline classifier but
compare our results against the HK classifier.

\begin{table}[htbp]
\small
\centering
\begin{tabular}{|l|c|}
\hline
Features & ACC \\\hline
Baseline & $77.4239$ \\
HK & $82.2976$ \\
1 & $81.8971$\\
2 & $83.3375$ \\
3 & $83.4291$ \\
4 & $83.5284$ \\
5 & $83.5393$ \\
6 & $83.5382$ \\
7 & $\mathbf{83.5682}$ \\\hline
\end{tabular}
\caption{Five-fold cross-validation accuracy for various lengths of $p$.}
\label{tab:cv}
\end{table}

Encouraged by this positive result, we proceeded to test if the combination of
HK features and subsequence features improve the cross-validation accuracies.
The
results of this experiment is shown in table~\ref{tab:cv_comb}. In this
experiment, the highest result is for $p=4$. We report the results for $p>1$
since, the $p=1$ classifier performs worse than HK-based classifier.
\begin{table}[h]
\small
\centering
\begin{tabular}{|l|c|}
\hline
Features & ACC \\\hline
HK+2 & $84.0117$ \\
HK+3 & $84.044$\\
HK+4 & $\mathbf{84.047}$\\
HK+5 & $84.0427$\\
HK+6 & $84.0448$\\
HK+7 & $84.027$\\\hline
\end{tabular}
\caption{Five-fold cross-validation accuracy for a combination of HK
features and various subsequence lengths.}
\label{tab:cv_comb}
\end{table}
These results show the superiority of subsequence features over word similarity
features. Now, we move on to test the performance of subsequence features
in a real-world scenario.

\subsubsection{Subfamily experiments}\label{subsub:sfmly}
The datasets used in these experiments are intended to imitate the real world
situation where there are gaps in knowledge regarding the cognate status of
word pairs.
%  A glance at the history of Indo-European scholarship reveals
% that there existed a large wealth of scholarship on few languages -- Latin,
% Greek, Sanskrit, and Gothic -- as against the remaining languages
% in the family. An accumulated scholarly work recognized hitherto unsuspected
% languages -- Armenian and Hittite -- as Indo-European and brought them into the
% family-folds.
In NLP, the default ratio between training and test datasets is
$4$:$1$ or $3$:$1$. In comparative linguistics, the amount of available labeled
data would
be much less. Hence, a $50$-$50$ random split of the
language groups tests the efficiency of subsequence features for cognate
identification and phylogenetic inference. We perform two sets of experiments
with the randomly split language groups.

% Write about development dataset.

The first set of experiments consist of testing the performance of subsequence
based features against HK features on multiple aspects. The results of this
experiment is given in table~\ref{tab:genus}. The results suggest that the
subsequence features perform consistently over $p \in [2,7]$. The $p=3$ based
linear classifier performs the best across all the evaluation measures. All the
subsequence based features agree on AP score and perform better than HK
classifier. The subsequence-based features outperform at MCC and ACC evaluation
measures.

\begin{table}[htbp]
\small
\centering
\begin{tabular}{|c|c|c|c|}
\hline
Features & ACC & MCC & AP \\ \hline
HK & $81.2034$ & $0.4269$ & $0.6565$\\
2 & $81.9048$ &  $0.4542$ & $0.6662$\\
3 & $\mathbf{82.0968}$ &  $\mathbf{0.4618}$ &
$\mathbf{0.6674}$\\
4 & $81.9451$ &  $0.4558$ & $0.6664$\\
5 & $81.9533$ &  $0.4561$ & $0.6665$\\
6 & $81.9117$ &  $0.4546$ & $0.6663$\\
7 & $81.9281$ &  $0.4552$ & $0.6664$\\ \hline
\end{tabular}
\caption{Performance of subsequence features on subfamily test set.}
\label{tab:genus}
\end{table}

We tested if the results of $p=3$ classifier is better than the HK classifier
using a \emph{paired t-test}. A classifier's agreement/disagreement ($1/0$)
with gold standard classification is encoded as a binary vector. Then, a paired
t-test is used to determine if there is a statistically significant difference
between the two classifiers. The difference between HK and $p=3$ is significant
at the $0.001$ level. Now we move to the results of the combination
experiment.

In this experiment, we use the same training and testing set but use the feature
combination explored in the cross-validation experiments. In
these experiments, the HK+2-based classifier won across all the evaluation
measures. The combination classifiers perform similarly on all the
evaluation measures. We ranked HK+2 classifier for the reason that the
classifier has lesser number of parameters and can be computed in lesser
time than the rest of the classifiers.
\begin{table}[htbp]
\small
\centering
\begin{tabular}{|l|c|c|c|}
\hline
Features & ACC & MCC & AP \\ \hline
HK+2 & $\mathbf{82.8121}$ &  $\mathbf{0.4877}$ &
$\mathbf{0.7017}$\\
HK+3 & $82.8039$ &  $0.4872$ & $0.7015$\\
HK+4 & $82.7655$ &  $0.4857$ & $0.7012$\\
HK+5 & $82.7674$ &  $0.4858$ & $0.7012$\\
HK+6 & $82.7649$ &  $0.4857$ & $0.7012$\\
HK+7 & $82.7655$ &  $0.4857$ & $0.7012$ \\ \hline
\end{tabular}
\caption{Performance of combination of subsequence and HK features on
subfamily test set.}
\label{tab:genus_comb}
\end{table}

A paired t-test shows that the difference between HK and HK+2 classifier's
predictions are significant at the $0.001$ level. Also, the difference between
the HK+2 and $p=3$ classifiers is significant at the $0.001$ level. We conclude
by observing that subsequence-based classifiers perform better than a HK-based
classifier.
% For each of the best classifiers, we look at the top $5$ concepts where the
% classifiers make mistakes in terms of FP and FN. We also checked if there is any
% relation between the number of cognate classes in a concept and the
% corresponding FPs and FNs through Pearsons's $r$. The correlation was always
% negative or close to zero suggesting that there is no relation between
% classifier errors and number of CCNs for a concept. The concept `mother'
% is at the top of list in FPs. In case of FNs, `who', `four', `five', and `how'
% are common in the top-5 list.

Now, we proceed to do an error analysis and then attempt to use our cognate
judgments for the purpose of phylogenetic
inference described in the next section.

\subsection{Error analysis}\label{subsec:errana}
In this section, we examine the misclassified word pairs. Our hypothesis is that
majority of FPs are correlates and FNs are those items which are quite
dissimilar. The gold standard cognate classification of a word pair is binary in
nature and cannot be used to measure the exact form similarity of a word pair.
In lieu, we use length normalized edit distance (LDN) to measure the difference.
To test our hypothesis about FNs and FPs, we correlated the classifier scores of
word pairs in each error class and classifier with the corresponding length
normalized edit distance scores. We expect a negative correlation between the
classifier scores and LDN scores since the former are similarity
scores. In fact, the correlations are negative as in
table~\ref{tab:err_corr}.
\begin{table}[h!]
\centering
\small
\begin{tabular}{|l|c|c|}
\hline
Classifier & FP & FN \\\hline
$p=3$ & $-0.29$ ($0.56$) & $-0.42$ ($0.38$)\\
HK+2 & $-0.55$ ($0.55$) & $-0.48$ ($0.38$)\\\hline
\end{tabular}
\caption{Correlation between probability scores and LDNs. The average of a
classifier's probabilities is shown in $(\dots)$.}
\label{tab:err_corr}
\end{table}

\section{Phylogenetic inference}\label{sec:phyli}
We describe a popular tree inference algorithm known as Neighbor-Joining
 (NJ) algorithm~\cite{saitou1987neighbor}. Then, we describe our gold standard
tree and Generalized Quartet distance (GQD) for measuring the distance between
the inferred tree and the gold standard tree.
\subsection{Tree inference}\label{sec:ti}
The cognate judgments returned by the linear classifier can be used to compute
the distance between a distance matrix, $D$, containing the distances between
all the language pairs in the test set. We can define binary and
similarity-based distance matrices from the classifier judgments. The binary
distance $d_{ij}^b$ between languages $i,j$ is defined as $1 - \frac{|\{k|
\hat{y_k} = 1\}|}{n_{ij}}$, where $n_{ij}$ is the total number of word pairs
between $i,j$. As mentioned earlier, the sigmoid function maps the linear
score of a classifier into $p(\hat{y_k}) \in [0,1]$. $p(\hat{y_k})$ can be
used to define the classifier distance $d_{ij}^s$ as $1 - \frac{\sum_{k}
p(\hat{y_k})}{n_{ij}}$.

The matrix $D$ is then supplied as an input to the NJ
algorithm\footnote{Available on \url{http://splitstree.org/}} to infer a tree
between the languages. The test set has $41$ languages and there are
$\approx10^{19}$ possible unrooted trees for $40$ languages. The problem of
exact tree
search is a computationally hard problem and there exist heuristic techniques to
reduce the searchable tree space.

NJ algorithm is a clustering algorithm which
has a complexity of $\mathcal{O}(l^3)$, $l$ is the number of languages, and is
shown to converge quickly for biological datasets consisting of thousands of
species.
% It is known to produce correct tree when the distance matrix
% is correct~\cite{felsenstein2004inferring}. It means that if we supply the
% right distance matrix which reflects the actual tree then, the algorithm does
% not make any mistake in finding the right tree from the large tree space.
NJ has been widely tested over both
real and simulated datasets and was reported to be statistically consistent
over different test conditions.
%  We supply $D$ as an input to the NJ algorithm
% for inferring the phylogenetic tree for the $41$ languages in the test set.

\subsection{Gold standard tree}\label{subsec:gst}

\begin{figure}[!ht]
\centering
\includegraphics
[height=0.9\textwidth,width=0.45\textwidth]{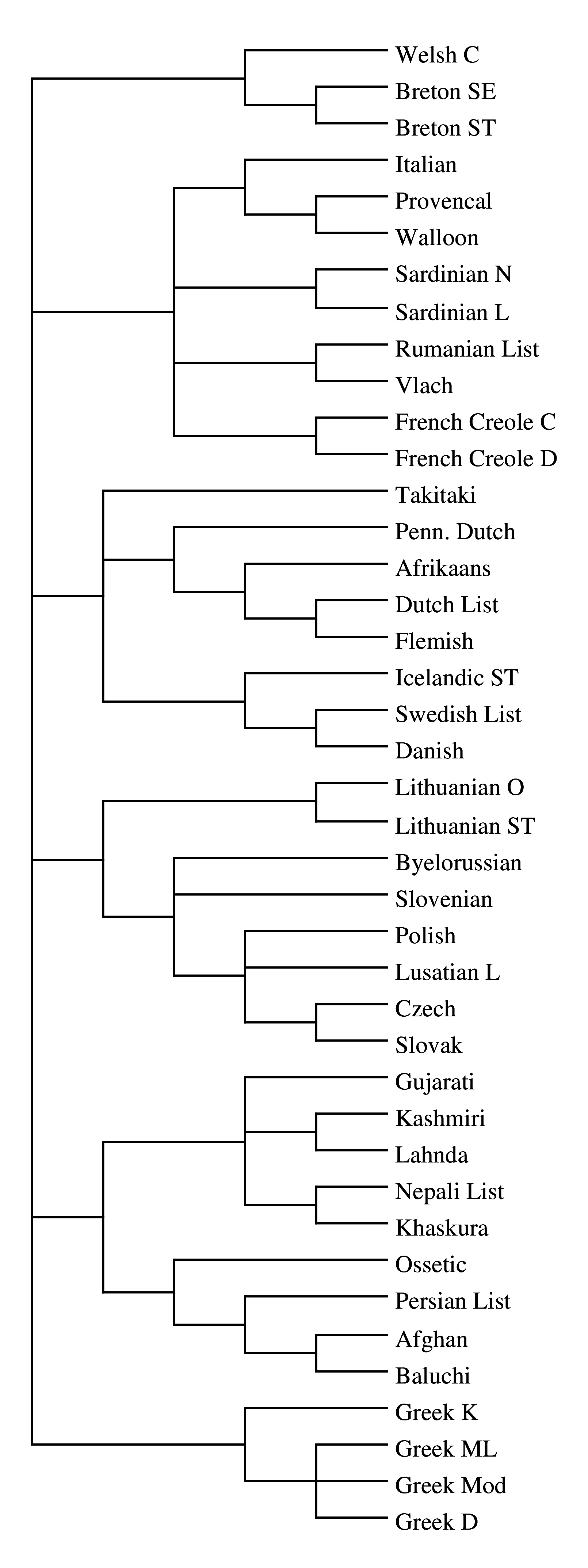}
\caption{The gold standard classification of the test set.}
\label{fig:fig1}
\end{figure}

The gold standard classification is shown in figure~\ref{fig:fig1}. Although,
Dyen et al., provide a classification for the languages in the test set, the
classification wrongly places the languages in the tree. Hence, we extract the
relevant languages from the expert classification given
in~\newcite{nordhoff2012glottolog}. The highest level of the tree is polytomous
or shows non-binary branching. The nature of highest level branching is still an
open question in Indo-European historical linguistics. Hence, our gold standard
tree also
shows the gaps in the scholarship. In fact, the tree
evaluation metric that will be introduced in the next subsection attempts to
alleviate this issue.

\subsection{Tree distance measure}\label{subsec:tsm}
In evolutionary biology, tree distance measures are used to measure the
accuracy of a tree inference algorithm. Quartet distance is the
state-of-the-art tree distance measure used to compute the distance between two
trees. Quartet distance is defined as the number of different quartets between
the trees. A quartet is a subtree with four leaves and there are ${l \choose
4}$ quartets in a tree with $l$ leaves.

A quartet is \emph{resolved} if there exists an
internal node that separates a pair of leaves. For example, the quartet
consisting of Swedish, Danish, Icelandic, and Dutch is resolved since Swedish
and Danish are separated from Icelandic and Dutch through an internal node.
Such a quartet is known as a \emph{butterfly} quartet. A star quartet
is complementary to a butterfly quartet since all the languages in a star
quartet are connected to a central node. The top node in the figure
\ref{fig:fig1} is an example of a star quartet.

The quartet distance (QD) between two
trees, $T_1, T_2$ is defined as:
\begin{equation}
\label{eq:qd}
% \small
\frac{\scriptstyle q(T_1)+q(T_2)-2s(T_1, T_2)-d(T_1, T_2)}{\scriptstyle{l
\choose 4}}
\end{equation}
\normalsize
where $q(T)$ is the number of butterflies in $T$, $s(T_1, T_2)$ is the number of
shared butterflies between $T_1, T_2$, and $d(T_1, T_2)$ is the number of
different butterflies between $T_1, T_2$.

\newcite{christiansen2006fast} developed
a fast algorithm for computing the quartet distance between trees having
thousands of leaves. The QD formula in equation \ref{eq:qd} counts the number
of resolved quartets in the inferred tree as errors. The inferred binary tree
$T_i$ should not be
penalized for the unresolvedness in the gold standard tree
$T_g$. \newcite{pompei2011accuracy} defined a new measure known as GQD to negate
the
effect of star quartets in $T_g$. GQD is defined as $d(T_i, T_g)/q(T_g)$. We use
both QD and GQD to evaluate the quality of the inferred trees.

\subsection{Tree inference results}
\begin{figure}[h]
\centering
\includegraphics
[height=0.7\textwidth,width=0.4\textwidth]{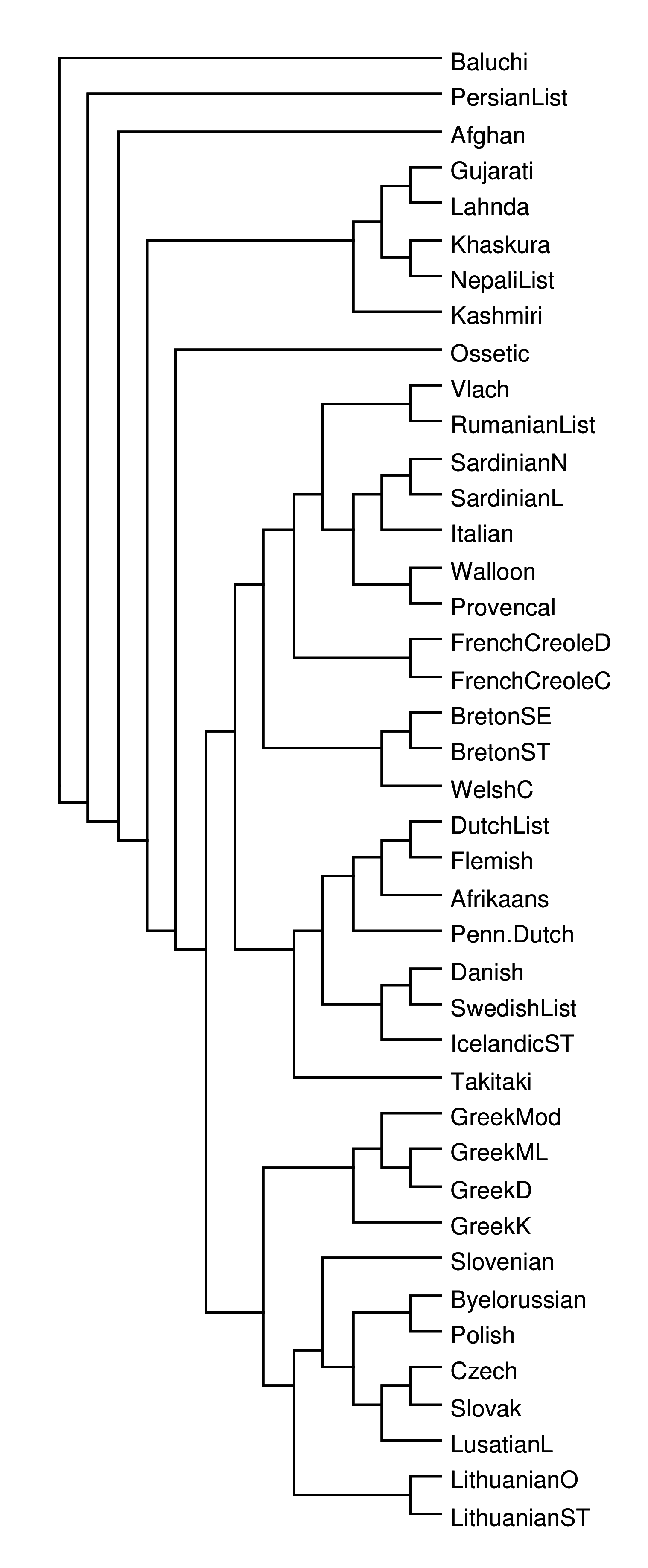}
\caption{The best tree on the test set based on HK (binary) classifier.}
\label{fig:fig2}
\end{figure}

\begin{table}[h]
\centering
\small
\begin{tabular}{|l|c|c|}
\hline
Classifier & QD & GQD \\\hline
Oracle & $0.29766$ & $0.005648$ \\
HK (wts.)& $0.30236$ & $0.012303$ \\
HK (binary)& $\mathbf{0.301669}$ & $\mathbf{0.011324}$ \\
$p=3$ (wts.) & $0.303782$ & $0.014316$ \\
$p=3$ (binary) & $0.304246$ & $0.014973$\\
HK+2 (wts.) & $0.30236$ & $0.012303$\\
HK+2 (binary) & $\mathit{0.30234}$ & $\mathit{0.012275}$\\
\hline
\end{tabular}
\caption{QD and GQD for the top performing models in
subfamily experiment.}
\label{tab:treedist}
\end{table}

In this experiment, we compute distance matrices from the predictions of the
top classifiers in section~\ref{subsub:sfmly}. The winning classifiers are
subsequence classifiers: $p=3$ and HK+2. We compare the winning classifiers
against HK classifier. The output of each classifier is used
to compute the binary and weighted (shown in table~\ref{tab:treedist} as
``(wts.)'') distance matrices based on equations defined above. In order to see
the effect of tree inference algorithm, we also report the difference
between the tree inferred from gold standard cognate judgments (\emph{Oracle
tree}) and the gold standard tree given by expert historical linguists.

Each phylogenetic tree is compared to the gold standard tree (cf.
figure~\ref{fig:fig1}) using the tree distance measures, QD and GQD. All the
classifiers give similar results in this experiment. The results suggest that
the choice of binary vs. weighted cognacy judgments do not make a significant
difference in the quality of the inferred trees. The results for HK-based
classifier are shown in bold-face since, it gives the best result and is also
the simplest of all the classifiers in terms of model complexity. The oracle
tree also differs from the expert classification.

The main difference between
the HK (binary) tree and the next best tree (italicized results in
table~\ref{tab:treedist}) is the placement
of Takitaki language. Both trees misplace Ossetic as an outlier in the
Indo-Iranian branch whereas, it should have been placed together with Iranian
branch. The HK+2 (binary) tree places Byelorussian correctly whereas HK
(binary) tree misplaces it. Italian's position is correctly determined in HK+2
(binary) tree whereas HK tree misplaces it. Overall, the difference between the
top-two trees is not large.
% The Oracle tree is quite close to the expert
% classification.

% Put the tree for the best classifier and tell where it makes mistakes.
% Write about language names.
\section{Conclusion and future work}\label{sec:concl}
In this paper, we introduced subsequences and tested their efficacy for cognate
identification and phylogenetic inference in a scenario where there is
incomplete knowledge about a language family. We showed that subsequences
perform significantly better than simple word similarity based classifiers for
cognate identification. We evaluated the performance of the classifiers at the
task of phylogenetic inference and found that there is no significant
difference between the various classifiers.

As a future work, we intend to employ fuzzy subsequence matching for building
the feature vectors for a word pair using a phonetic similarity measure. We also
intend to integrate articulatory features of sounds into our experiments. We
plan to test our features on Austronesian vocabulary
lists~\cite{greenhill2008austronesian}. Further, we plan to test the subsequence
features for automated classification of thousands of languages available in
ASJP database~\cite{wichmann2010evaluating}.

\section*{Acknowledgments}
I warmly thank Richard Johansson, Johann-Mattis List, and Søren Wichmann for all 
the comments
which made the draft better. The paper was originally submitted to EMNLP 2014 
but was rejected. I benifitted 
substantially from the comments made by all the three people.
% How the best subsequence values explain linguistic reality?
% 
% Use inexact matching of subsequences.
% 
% Use factored n-grams for extracting features.
% 
% Use a phonetic similarity between subsequences.
% 
% Use of skip-grams in IR. Work of järvelin relate to cognate identification
% task. Relation between cognate identification task and IE/IR task.

% \section*{Acknowledgments}

% The acknowledgments should go immediately before the references.  Do
% not number the acknowledgments section. Do not include this section
% when submitting your paper for review.

\bibliographystyle{acl}
\bibliography{myreflnks}

\end{document}